\documentclass[conference]{IEEEtran}

\usepackage{cite}
\usepackage{amsmath,amssymb,amsfonts}
\usepackage{graphicx}
\usepackage{textcomp}
\usepackage{xcolor}
\usepackage{booktabs}
\usepackage{tabularx}
\usepackage{float} 
\usepackage[caption=false,font=footnotesize]{subfig}

\usepackage[utf8]{inputenc}
\usepackage{arabtex}
\usepackage{utf8}
\setcode{utf8}

\graphicspath{{paper_figs/}}

\usepackage{hyperref}
\hypersetup{breaklinks=true}
\title{Saudi-Dialect-ALLaM: LoRA Fine-Tuning for Dialectal Arabic Generation}

\author{
\IEEEauthorblockN{Hassan Barmandah}
\IEEEauthorblockA{
Department of Software Engineering\\
Umm Al-Qura University, Makkah, Saudi Arabia\\
s445001043@uqu.edu.sa}
}

\begin{document}
\emergencystretch=3em
\maketitle

\begin{abstract}
Large language models (LLMs) for Arabic are still dominated by Modern Standard Arabic (MSA), with limited support for Saudi dialects such as Najdi and Hijazi. This underrepresentation hinders their ability to capture authentic dialectal variation. Using a privately curated Saudi Dialect Instruction dataset (Hijazi \& Najdi; 5{,}466 synthetic instruction--response pairs; 50/50 split), we LoRA-tune ALLaM-7B-Instruct-preview—the first foundation model developed in Saudi Arabia—for Saudi dialect generation. We investigate two variants: (i) \textit{Dialect-Token} training, which prepends an explicit dialect tag to the instruction, and (ii) \textit{No-Token} training, which omits the tag at formatting time. Evaluation on a held-out test set combines an external dialect classifier with text fidelity metrics (chrF++ and BERTScore) and diversity measures. The Dialect-Token model achieves the best control, raising the Saudi rate from 47.97\% to 84.21\% and reducing MSA leakage from 32.63\% to 6.21\%; fidelity also improves (chrF++ $+3.53$, BERTScore $+0.059$). Both LoRA variants outperform strong generic instruction models (Falcon-7B-Instruct, Llama-3.1-8B-Instruct, Qwen-2.5-7B-Instruct, AceGPT-v2-8B-Chat, JAIS-13B-Chat) in dialect control and fidelity, while avoiding metadata-tag echoing that these baselines frequently exhibit. \textbf{We do not release the dataset or any model weights/adapters}; instead, we release training/evaluation/inference code and a detailed datasheet (schema and aggregate statistics) to support independent verification.
\end{abstract}


\section{Introduction}

Large language models (LLMs) such as LLaMA~\cite{llama3modelcard} and Arabic-centric systems like ALLaM~\cite{bari2025allam}, JAIS-13B-Chat~\cite{sengupta2023jais}, and AraGPT2~\cite{antoun-etal-2021-aragpt2} have accelerated progress in Arabic NLP. However, coverage remains skewed toward \textit{Modern Standard Arabic} (MSA), with comparatively limited support for regional dialects. This imbalance often yields overly formal, pan-Arabic outputs that fail to capture cultural and pragmatic nuance. In particular, Saudi dialects—Najdi and Hijazi—are underrepresented in open models despite their widespread real-world use.

Insufficient support for Saudi dialects has practical consequences. When systems default to MSA, they underperform in everyday applications such as dialogue, education, entertainment, and culturally aware assistants. Community leaderboards (e.g., OALL)~\cite{OALL-2} repeatedly highlight gaps in dialectal coverage. Moreover, widely used resources such as MARBERT reveal that Gulf Arabic versus MSA remains a confusable axis~\cite{abdulmageed2021_marbert}. In our evaluation, we therefore adopt the MARBERTv2 Arabic Written Dialect Classifier~\cite{ibrahimamin_marbertv2_arabic_written_dialect_classifier}, a five-way model (MAGHREB/LEV/MSA/GLF/EGY) fine-tuned from MARBERTv2 for written dialect identification, where GLF serves as a practical proxy for Saudi usage.

\textbf{Baselines.} We include Falcon-7B-Instruct~\cite{falcon40b} as a baseline. While TII has announced Falcon-Arabic with strong results on Arabic benchmarks (e.g., OALL v2)~\cite{falcon-arabic, OALL-2}, we evaluate publicly available Falcon baselines and discuss Falcon-Arabic qualitatively. Alongside Falcon, we compare against Llama-3.1-8B-Instruct~\cite{llama3modelcard}, Qwen-2.5-7B-Instruct~\cite{qwen2.5}, AceGPT-v2-8B-Chat~\cite{liang2024alignment}, and JAIS-13B-Chat~\cite{sengupta2023jais}.

\textbf{This work.} We curate a balanced, synthetic instruction–response dataset for Saudi dialects (Hijazi/Najdi; 5{,}466 pairs) and train a LoRA-tuned variant of ALLaM-7B-Instruct-preview specialized for Saudi dialect generation. We study two strategies: (i) \textit{Dialect-Token Conditioning}, which prepends an explicit dialect tag to instructions, and (ii) \textit{No-Token Conditioning}, which omits tags and relies on the model to infer dialect implicitly. While the dataset and trained weights/adapters are not publicly released, we provide a comprehensive datasheet (schema, cleaning pipeline, topic taxonomy, and descriptive statistics) together with full training/evaluation/inference code to support reproducibility.

On a held-out Saudi test set, the Dialect-Token variant achieves higher dialect fidelity (84.2\% Saudi per the external classifier~\cite{ibrahimamin_marbertv2_arabic_written_dialect_classifier}) while substantially reducing MSA leakage (6.2\%). Both LoRA variants outperform strong generic instruction models (Falcon-7B-Instruct, Llama-3.1-8B-Instruct, Qwen-2.5-7B-Instruct, AceGPT-v2-8B-Chat, JAIS-13B-Chat) on Saudi-specific automatic metrics, while avoiding issues such as tag echoing observed in those baselines.

\textbf{Contributions.} Our work makes four contributions: \\
(1) a curated, balanced instruction–response dataset dedicated to Saudi Arabic (Hijazi/Najdi); although private, we disclose a detailed datasheet and aggregate statistics; \\
(2) an experimental study and open codebase for LoRA adaptation of ALLaM-7B-Instruct-preview for Saudi dialect generation; neither the dataset nor trained weights/adapters are released; \\
(3) a systematic comparison of Dialect-Token vs.\ No-Token strategies, demonstrating the advantages of explicit conditioning for dialect control and MSA-leakage reduction; \\
(4) a transparent evaluation suite combining an external dialect classifier with fidelity and diversity metrics to enable independent verification without access to the raw training set or model weights.

\section{Related Work}

Research on Arabic LLMs and dialect modeling has expanded rapidly with the rise of instruction tuning and Arabic-native evaluation. A recurring challenge is the dominance of Modern Standard Arabic (MSA) in pretraining corpora, which leaves regional varieties—especially Gulf and Saudi dialects—underrepresented. Below, we review seven lines of work shaping today’s Arabic LLM landscape across pretraining, instruction data, and dialect identification, framing our focus on Saudi (Hijazi/Najdi) instruction tuning and GLF-based external evaluation.

Bari et al.\ \cite{bari2025allam} introduce \textit{ALLaM-7B-Instruct-preview}, the first Saudi foundation model~\cite{humaine_huwaimain2025}. Their work covers Arabic-centric pretraining and instruction-tuned variants, evaluated through automatic metrics, LLM-as-a-judge assessments, and human studies. It highlights practical recipes for aligning Arabic models and demonstrates strong performance across general Arabic tasks. Since our backbone is ALLaM-7B-Instruct-preview, this research provides the architectural and training substrate on which we apply LoRA and Saudi-dialect supervision, illustrating how targeted data can further specialize a Saudi foundation model for dialect-specific generation.

Qarah \cite{qarah2024saudibert} proposes \textit{SaudiBERT}, an encoder pre-trained on large-scale Saudi corpora (e.g., STMC tweets and Saudi forums). The study argues that dialect-specific pretraining better captures Saudi lexical, morphological, and pragmatic phenomena than MSA-centric encoders. This strengthens the case for Saudi-centric resources; we extend this idea to the generative setting by curating (private) a balanced Saudi instruction dataset and measuring dialectal fidelity in generation.

Abdul-Mageed et al.\ \cite{abdul-mageed-etal-2021-arbert} present \textit{ARBERT/MARBERT}, Arabic transformers trained on massive social media text covering diverse dialects. Beyond improving many Arabic benchmarks, these models underpin later dialect-aware tools and datasets. Their emphasis on informal, user-generated content foreshadows our evaluation focus: distinguishing Gulf/Saudi-style outputs from MSA in open-ended generation.

Amin \cite{ibrahimamin_marbertv2_arabic_written_dialect_classifier} releases the \textit{MARBERTv2 Arabic Written Dialect Classifier}, a five-way identifier (GLF/LEV/MSA/EGY/MAGH) fine-tuned from MARBERTv2~\cite{abdul-mageed-etal-2021-arbert} for short, written text. Community usage has converged on this classifier as a practical external judge for dialect labels in generated text. We adopt it to quantify Saudi/Gulf alignment (via GLF) and to estimate MSA leakage, providing an objective complement to human judgments.

Sengupta et al.\ \cite{sengupta2023jais} introduce \textit{JAIS-13B-Chat}, Arabic-centric foundation and instruction-tuned models trained on mixed Arabic–English corpora. The paper shows that Arabic-first post-training substantially boosts instruction following in Arabic while retaining multilingual utility. We situate our LoRA-tuned Saudi models against such open baselines to test whether targeted Saudi supervision improves dialect fidelity beyond generic instruction tuning.

The Falcon team \cite{falcon-arabic} recently introduced \textit{Falcon-Arabic}, a 7B-parameter model adapted from Falcon~3. Unlike earlier efforts that relied on translated data, Falcon-Arabic was trained on high-quality native Arabic corpora and extended with 32k Arabic-specific tokens to better capture morphology and dialectal variation. The model excels in MSA while also demonstrating strong coverage of regional dialects, outperforming not only Arabic-focused models of similar size but even larger multilingual systems across benchmarks such as Arabic MMLU, MadinahQA, and Aratrust. With a context length of 32k tokens and alignment via supervised fine-tuning and Direct Preference Optimization, Falcon-Arabic sets a new bar for Arabic-first LLMs. Our work differs in focus: while Falcon-Arabic provides broad coverage across MSA and multiple dialects, \emph{we curate (private)} a fine-grained Saudi-centric dataset and train a LoRA-tuned variant of ALLaM-7B-Instruct-preview (weights not released), targeting precise control of Hijazi and Najdi generation.

Chouikhi et al.\ \cite{gemmar_arxiv_2024} introduce \textit{GemmAr}, which leverages large-scale Arabic instruction data. Its companion resource, InstAr-500k~\cite{chouikhi2024}, demonstrates that broad Arabic-focused instruction tuning yields sizable gains even for models not originally trained for Arabic. This underscores the leverage of instruction data quality and coverage. Our contribution complements this by balancing dialect labels and testing explicit dialect-token conditioning to control Saudi style during generation.

Taken together, the reviewed literature underscores a shift from MSA-centric modeling toward dialect-aware Arabic NLP, spanning pretraining (e.g., SaudiBERT), instruction tuning (e.g., ALLaM, JAIS, GemmAr/InstAr-500k), and standardized dialect identification via MARBERT-family tools. Building on these advances, our study targets the underrepresented Saudi varieties (Hijazi/Najdi) \emph{by curating} a balanced instruction–response dataset (kept private) and applying parameter-efficient fine-tuning of ALLaM-7B-Instruct-preview with explicit dialect-token conditioning. Using an external written-dialect classifier (GLF/LEV/MSA/EGY/MAGH) as an objective judge, we quantify Saudi/Gulf alignment and MSA leakage, demonstrating consistent gains over strong open baselines and offering a reproducible path for culturally grounded, Saudi-centric generation.

\section{Dataset and Analysis}

\subsection{Dataset Description}
We curate the \textit{Saudi Dialect Instruction} dataset (Hijazi \& Najdi), a synthetic instruction–response corpus targeting Saudi Arabic. Each record consists of an \texttt{instruction}, a dialect-pure \texttt{response}, and a categorical \texttt{dialect} label. The dataset is \textbf{not publicly released}. We disclose the full schema, preprocessing steps, and descriptive statistics to support transparency.

\subsection{Distribution and Coverage}

To validate balance and topical coverage, we analyze dialect counts and category frequencies.

\begin{figure}[t]
  \centering
  \includegraphics[width=0.75\linewidth]{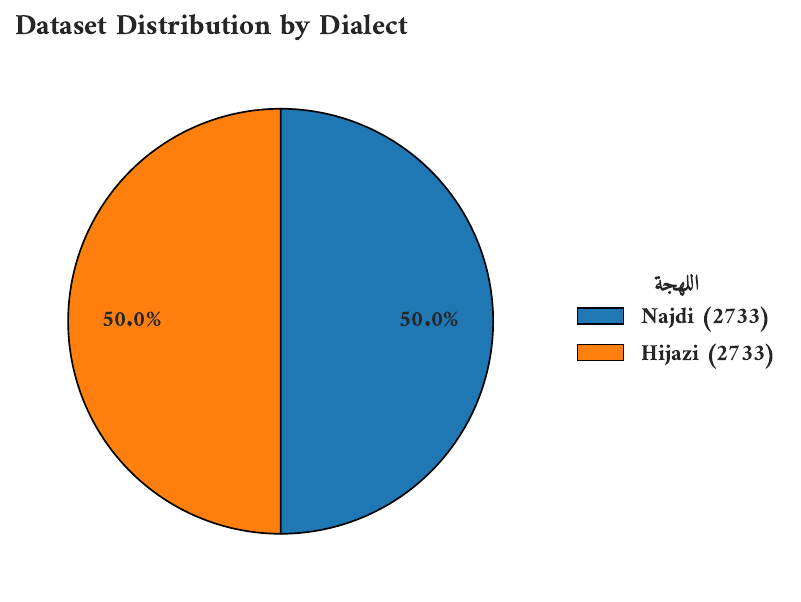}
  \caption{Dataset distribution by dialect (Hijazi vs.\ Najdi).}
  \label{fig:dialect-pie}
\end{figure}

Figure~\ref{fig:dialect-pie} confirms a strict 50/50 balance across Hijazi and Najdi labels. This design prevents dialect–topic confounds and ensures that evaluation performance cannot be attributed to skewed representation.

\begin{figure}[t]
  \centering
  \includegraphics[width=\linewidth]{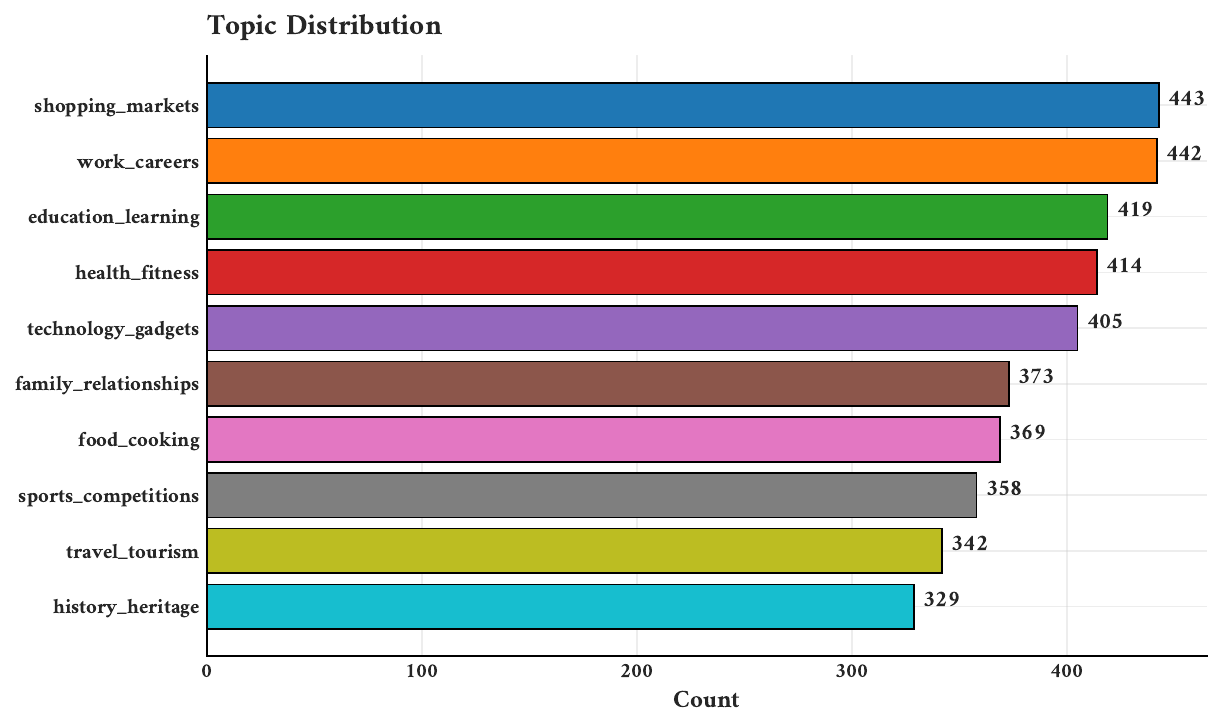}
  \caption{Top-10 topic frequencies in the dataset.}
  \label{fig:topic-bar}
\end{figure}

The most common topics are (shopping\_markets), (work\_careers), (education\_learning), (health\_fitness), and (technology\_gadgets). Together, these five domains account for roughly 55\% of the dataset, reflecting strong coverage of everyday contexts.

\subsection{Lexical Analysis}

We further analyze word-level distributions to inspect dialect authenticity.

\begin{figure}[t]
  \centering
  \includegraphics[width=\linewidth]{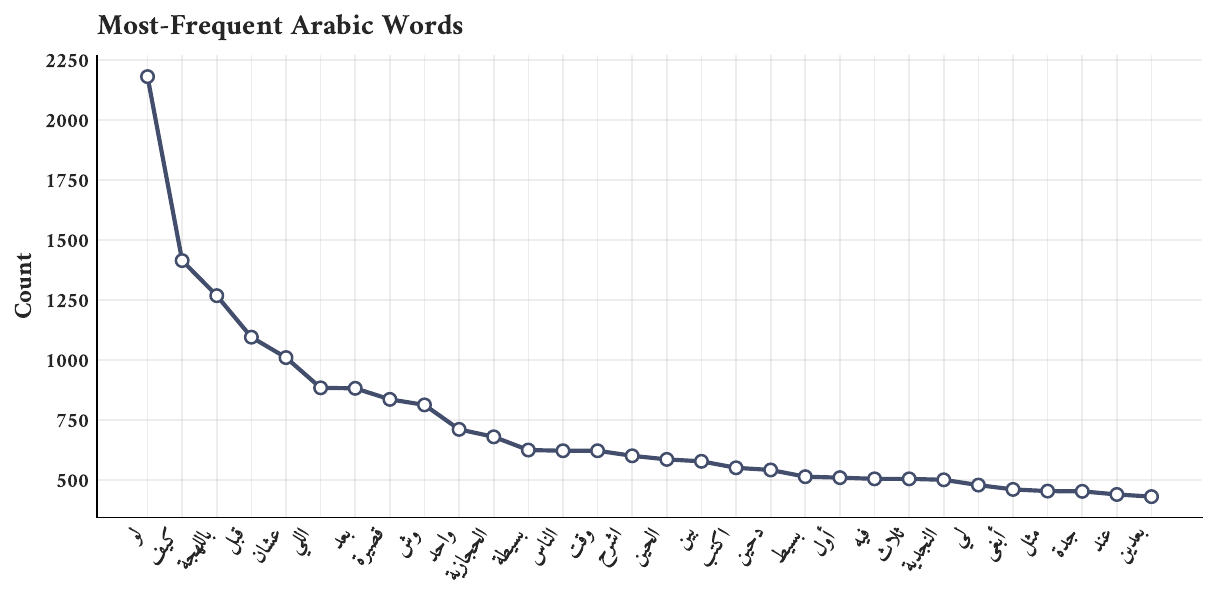}
  \caption{Ranked frequency distribution of Arabic tokens.}
  \label{fig:ranked-words}
\end{figure}

The ranked distribution highlights frequent Saudi dialect terms absent in MSA, confirming dialectal grounding. As a qualitative complement, we generate a word cloud of the most frequent tokens:

\begin{figure}[t]
  \centering
  \includegraphics[width=\linewidth]{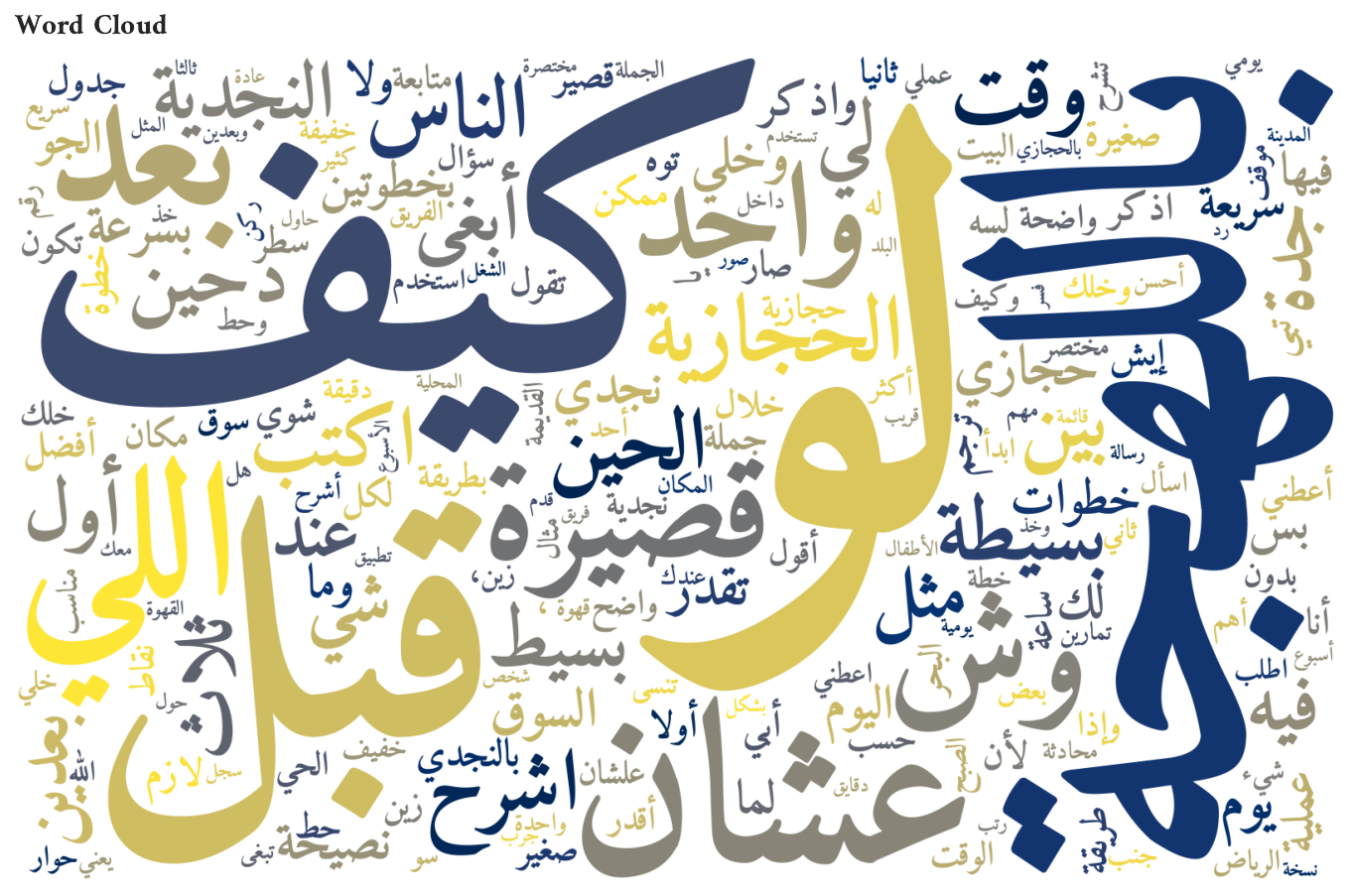}
  \caption{Word cloud (Arabic-only) across Hijazi and Najdi subsets.}
  \label{fig:wordcloud}
\end{figure}

The cloud reveals salient lexical items typical of Saudi everyday conversation, reinforcing dataset fidelity to regional speech norms.

\subsection{Data Preparation Pipeline}

To construct training-ready inputs, we apply two key steps:
\begin{itemize}
\item \textbf{Balance:} Enforce an exact 50/50 split between Hijazi and Najdi examples, ensuring equal representation while dropping any unknown or ambiguous dialect entries to maintain dataset integrity.  
\item \textbf{Dialect tags:} Prepend an explicit dialect token, such as \texttt{<DIALECT=HIJAZI>} or \texttt{<DIALECT=NAJDI>}, to the \texttt{instruction} field. This step enables controlled conditioning for generation and allows the model to clearly distinguish between Hijazi and Najdi outputs during training.
\end{itemize}

This pipeline supports both the Dialect-Token and No-Token strategies explored in Section~4.

\subsection{Limitations}

Despite strict cleaning and balancing, the dataset remains synthetic in nature and does not include multi-turn dialogue, natural conversations, or spontaneous interaction patterns. This limits its ability to capture discourse flow, speaker variation, and pragmatic context. Furthermore, the relatively small size of 5.5k instruction–response pairs restricts coverage of less common dialectal phenomena, idiomatic expressions, and low-frequency lexical items that occur in everyday communication.

Other Saudi-dialect corpora exist, such as SauDial (game localization)~\cite{aburayyash2025saudial}, SADLyC (song lyrics)~\cite{alahmari-2025-sadslyc}, and SADA (speech transcripts)~\cite{sada2022}. However, these differ in format, domain, and intended use cases. By contrast, our contribution is among the first balanced instruction–response corpora targeting Saudi Arabic (Hijazi and Najdi); however, the dataset is private and not shared publicly.

\section{Methodology}

\subsection{Data Preprocessing}
To ensure fair and balanced evaluation, the Saudi Dialect Instruction Dataset was split into train/dev/test sets with an 80/10/10 ratio. Stratified sampling was applied across three metadata dimensions—dialect (Hijazi, Najdi), topic (18 categories), and length (short, medium, long)—to preserve representative distributions in each split.

Each example was converted into the instruction–response format used for instruction tuning:
\begin{quote}
\texttt{Instruction: [user prompt]} \\
\texttt{Response: [dialectal output]}
\end{quote}

Text was tokenized using the ALLaM-7B-Instruct-preview tokenizer, with sequences truncated or packed to a maximum of 2048 tokens for GPU efficiency.  
We prepared two variants:
\begin{itemize}
    \item \textbf{Dialect-Token:} prepend an explicit tag (\texttt{<HIJAZI>} or \texttt{<NAJDI>}) to the instruction, giving the model direct conditioning signals;  
    \item \textbf{No-Token:} omit tags, leaving dialect inference implicit.  
\end{itemize}

This preprocessing pipeline yields a clean, balanced dataset suitable for LoRA fine-tuning of ALLaM-7B-Instruct-preview.

\subsection{Model Architecture}
We adopt ALLaM-7B-Instruct-preview~\cite{bari2025allam}, a 7B-parameter decoder-only transformer~\cite{vaswani2017attention} designed for large-scale language modeling. The model follows an autoregressive paradigm, where the probability of each token is conditioned on all preceding tokens. Input text is mapped into embeddings with positional encodings, then passed through stacked transformer blocks comprising multi-head self-attention and feedforward sublayers with residual connections and normalization. This layered design captures both local and long-range dependencies, enabling dialect-aware generation.

\noindent\textit{Scaled dot-product attention:}
\[
\text{Attention}(Q,K,V)=\mathrm{softmax}\!\left(\frac{QK^{\top}}{\sqrt{d_k}}\right)V
\]

\noindent\textit{Transformer block update (residual + norm + FFN):}
\[
\begin{aligned}
Z_l &= \mathrm{LayerNorm}\!\big(X_l+\text{Attention}(Q_l,K_l,V_l)\big), \\
X_{l+1} &= \mathrm{LayerNorm}\!\big(Z_l+\mathrm{FFN}(Z_l)\big).
\end{aligned}
\]

Here, $Q$, $K$, and $V$ represent query, key, and value projections, while $d_k$ denotes the key dimension. Attention provides contextual weighting of tokens, and residual–normalization ensures stable gradients. Stacking these blocks yields progressively richer contextual features.

\begin{figure}[t]
    \centering
    \includegraphics[width=\linewidth]{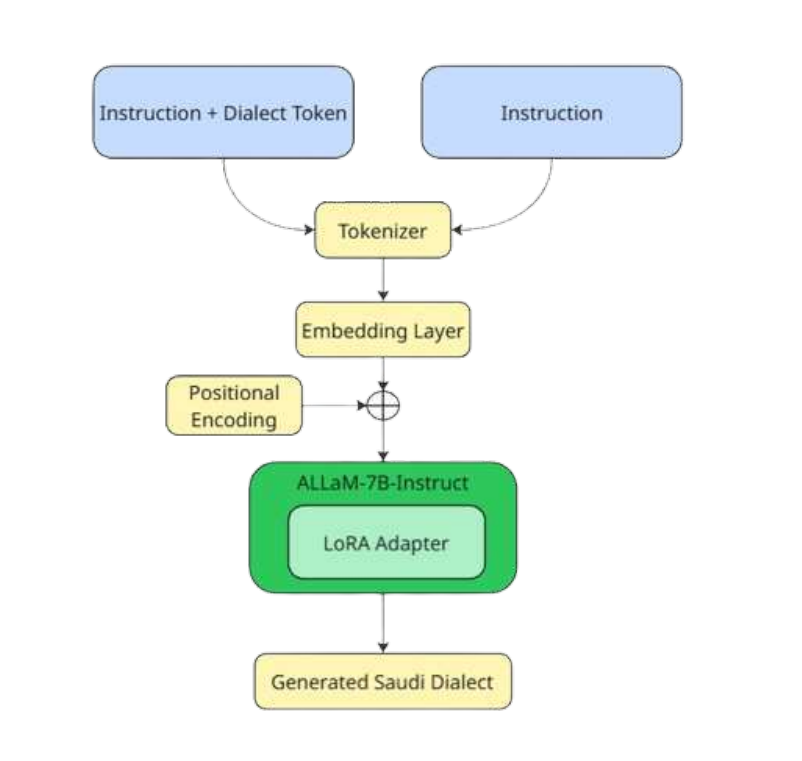}
    \caption{High-level pipeline of ALLaM-7B-Instruct-preview with LoRA adapters for Saudi dialect generation.}
    \label{fig:allam-arch}
\end{figure}

\subsection{Fine-Tuning}
We fine-tune ALLaM-7B-Instruct-preview using supervised next-token prediction with causal language modeling (CLM). For an instruction–response pair, the model minimizes cross-entropy between predicted token distributions and gold targets. Model perplexity (PPL) is tracked as an intrinsic measure of fit.

\noindent\textit{Cross-entropy loss:}
\[
\mathcal{L}_{CE} = - \frac{1}{N}\sum_{i=1}^N \log p_\theta(y_i \mid y_{<i}, x)
\]

Training ran for 15 epochs with batch size 2 (gradient accumulation of 8), cosine learning rate with warmup, maximum sequence length 2048, mixed-precision BF16, and sequence packing for efficiency.

\subsection{Parameter-Efficient Adaptation (LoRA)}
To adapt the ALLaM-7B-Instruct-preview backbone efficiently, we employ Low-Rank Adaptation (LoRA)~\cite{hu2021lora}. LoRA injects trainable low-rank matrices into projection layers, updating only these factors while keeping backbone weights frozen. This reduces memory and compute costs, making it practical for dialect specialization.

\noindent LoRA hyperparameters: 
\begin{itemize}
    \item Rank $r = 32$  
    \item Scaling factor $\alpha = 64$  
    \item Dropout $p = 0.1$  
    \item Bias: none  
    \item Task type: Causal LM  
\end{itemize}

Both Dialect-Token and No-Token models were trained with identical configurations, differing only in whether the leading dialect tag was included.

\subsection{Training and Inference Setup}

\noindent\textbf{Optimization.}  
Fine-tuning used AdamW with a cosine learning-rate schedule and restarts over 15 epochs. The effective batch size was 16 (per-device 2, gradient accumulation 8). Mixed-precision BF16 and gradient checkpointing reduced memory usage and improved efficiency.  

\noindent\textbf{Inference.}  
At generation time, we applied nucleus sampling~\cite{holtzman2020curious} with temperature $T=0.6$, top-$p=0.95$, and maximum length of 256 tokens, balancing fluency and diversity while avoiding excessive randomness.

\subsection{Evaluation Metrics}
We evaluate baselines and fine-tuned models with a multi-dimensional suite covering dialect classification, fluency, semantic similarity, and diversity:

\begin{itemize}
  \item \textbf{Saudi\% (GLF)} $\uparrow$ — proportion of outputs labeled Gulf Arabic (GLF) by the MARBERTv2 Written Dialect Classifier~\cite{ibrahimamin_marbertv2_arabic_written_dialect_classifier}. Higher = more faithful Saudi style.
  \[
    \text{Saudi\%} = \frac{1}{N} \sum_{i=1}^{N} \mathbf{1}\{\hat{y}_i = \text{GLF}\} \times 100
  \]
  
  \item \textbf{MSA leak\%} $\downarrow$ — proportion classified as MSA. Lower = less leakage into formal MSA.
  \[
    \text{MSA leak\%} = \frac{1}{N} \sum_{i=1}^{N} p(\text{MSA} \mid x_i) \times 100
  \]

  \item \textbf{Low-conf\%} $\downarrow$ — fraction of outputs with classifier confidence $<0.55$. Lower = more stable dialect assignment.
  \[
    \text{Low-conf\%} = \frac{1}{N} \sum_{i=1}^{N} \mathbf{1}\left\{\max_{c} p(c \mid x_i) < 0.55\right\} \times 100
  \]

  \item \textbf{chrF++} $\uparrow$ — character-level $n$-gram F-score measuring surface overlap between generated output $h_i$ and reference $r_i$~\cite{popovic2017chrfpp}.
  \[
    \text{chrF++} = \frac{2 \cdot P_{\text{char}} \cdot R_{\text{char}}}{P_{\text{char}} + R_{\text{char}}}
  \]

  \item \textbf{BERTScore F1} $\uparrow$ — semantic similarity using contextual embeddings~\cite{zhang2020bertscore}.
  \[
    \text{BERTScore} = \frac{2 \cdot P_{\text{BERT}} \cdot R_{\text{BERT}}}{P_{\text{BERT}} + R_{\text{BERT}}}
  \]

  \item \textbf{distinct-2/3} $\uparrow$ — lexical diversity, ratio of unique bigrams/trigrams to total~\cite{li2016diversity}.
  \[
    \text{distinct-}n = \frac{|\text{unique } n\text{-grams}|}{|\text{total } n\text{-grams}|}
  \]

  \item \textbf{Self-BLEU} $\downarrow$ — BLEU of each output $h_i$ against all others. Lower = higher variety~\cite{zhu2018texygen}.
  \[
    \text{Self-BLEU} = \frac{1}{N} \sum_{i=1}^{N} \text{BLEU}(h_i, \{h_j: j \neq i\})
  \]
\end{itemize}

All models, both external baselines and fine-tuned variants, are evaluated on the same held-out test set under this unified pipeline.

\section{Experimental Results}

We evaluate ALLaM-7B-Instruct-preview fine-tuned under two regimes (Dialect-Token vs.\ No-Token) and compare them against strong Arabic LLM baselines. Results are reported using both automatic metrics (dialect fidelity, text quality, and diversity) and human evaluation.

\subsection{Overall Performance}
The \texttt{ALLaM-LoRA-Token} model achieves the strongest overall performance, producing \textbf{84.2\%} Saudi-aligned generations while limiting MSA leakage to \textbf{6.3\%}. This demonstrates the benefit of explicit dialect conditioning. In comparison, the No-Token variant achieves 80.5\% Saudi correctness but exhibits slightly higher MSA interference.  

Human evaluation corroborates these findings: the token-based model received the highest ratings for dialect correctness (68.83\%) and fluency (74.83\%), consistent with automatic metric trends.

\subsection{Training and Validation Performance}
\begin{figure}[H]
\centering
\subfloat[Training loss]{\includegraphics[width=0.48\linewidth]{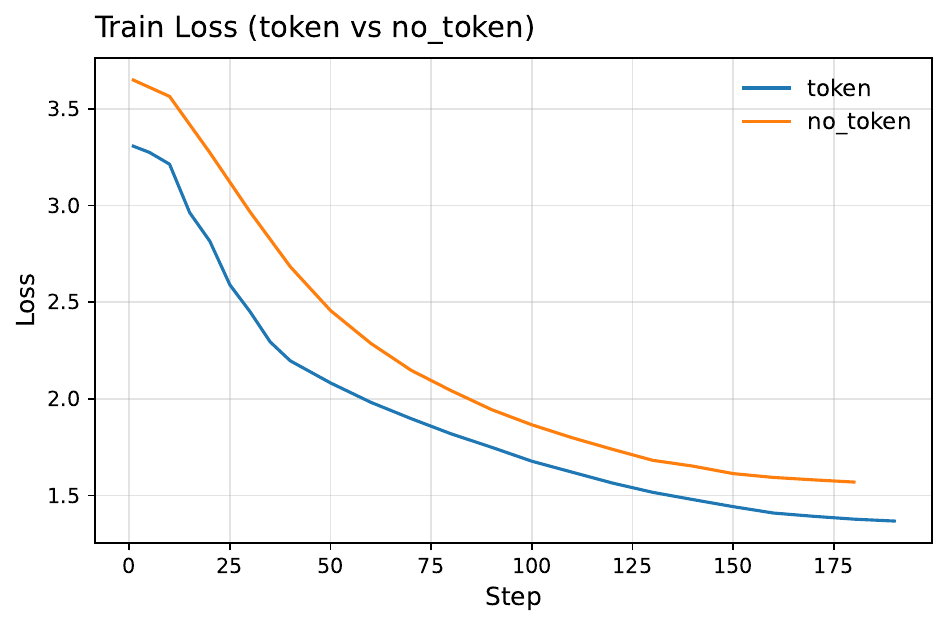}
\label{fig:train-loss}}
\hfill
\subfloat[Evaluation loss]{\includegraphics[width=0.48\linewidth]{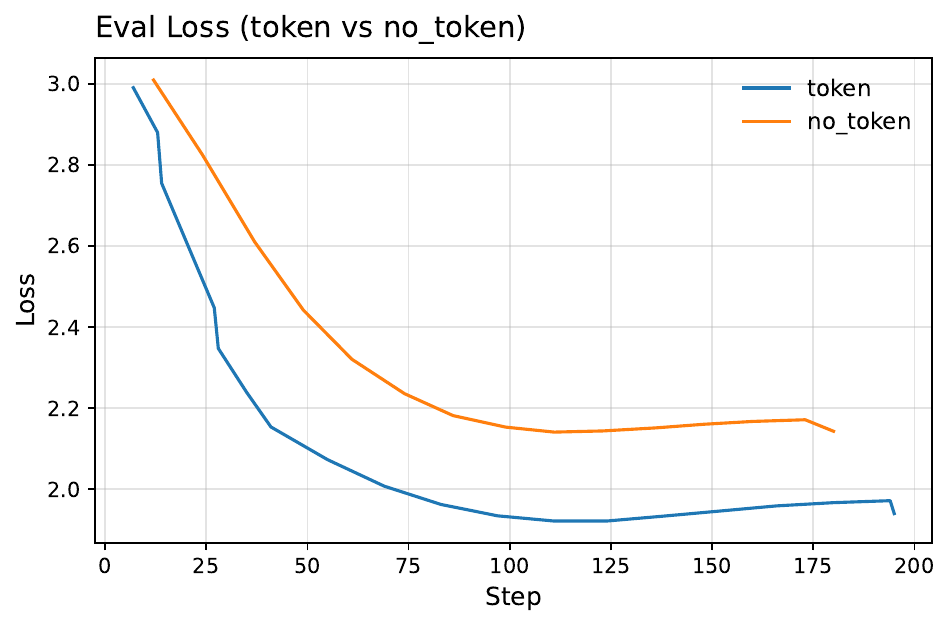}
\label{fig:eval-loss}}
\caption{Training and evaluation loss curves across epochs for Token vs.\ No-Token models.}
\label{fig:loss-combined}
\end{figure}

Figure~\ref{fig:loss-combined} shows training and evaluation loss across epochs.  
The token-based model converges faster and maintains lower validation loss than the No-Token variant, underscoring the effectiveness of explicit dialect supervision.

\subsection{Human Evaluation}
To complement automatic metrics, we conducted a human study with native Saudi speakers.  
Each annotator rated 40 prompts per model on a 1--5 Likert scale across three criteria:  
(1) \textit{Dialect Correctness},  
(2) \textit{Fluency / Naturalness}, and  
(3) \textit{Task Adherence}.  

Scores were normalized to percentages (5 $\mapsto$ 100\%). Table~\ref{tab:human-eval-percent} shows that \texttt{ALLaM-LoRA-Token} leads on both Dialect Correctness and Fluency while maintaining strong Task Adherence.

\begin{table}[H]
\caption{Human evaluation results (\%); higher is better.}
\label{tab:human-eval-percent}
\centering
\setlength{\tabcolsep}{4pt}
\begin{tabularx}{\columnwidth}{@{}>{\raggedright\arraybackslash}X r r r@{}}
\toprule
\textbf{Model} & \textbf{Dialect \%} $\uparrow$ & \textbf{Fluency \%} $\uparrow$ & \textbf{Task \%} $\uparrow$ \\
\midrule
\textbf{ALLaM-LoRA-Token} & \textbf{68.83} & \textbf{74.83} & \textbf{91.50} \\
ALLaM-LoRA-No-Token      & 66.92 & 72.67 & 88.50 \\
AceGPT-v2-8B-Chat        & 28.67 & 31.50 & 90.17 \\
Llama-3.1-8B-Instruct    & 28.17 & 31.25 & 72.92 \\
\bottomrule
\end{tabularx}
\end{table}

\begin{table*}[!t]
\centering
\caption{Automatic evaluation of dialect fidelity and diversity. (\textbf{Higher} is better for Saudi\%, chrF++, BERTScore, distinct-2/3; \textbf{Lower} is better for MSA leak, Low-conf, and Self-BLEU). Self-BLEU is reported on the 0--1 scale.}
\label{tab:dialect}
\resizebox{\textwidth}{!}{%
\begin{tabular}{@{}lrrrrrrrr@{}}
\toprule
Model & Saudi (\%) $\uparrow$ & MSA leak (\%) $\downarrow$ & Low-conf (\%) $\downarrow$ &
chrF++ $\uparrow$ & BERTScore F1 $\uparrow$ &
distinct-2 $\uparrow$ & distinct-3 $\uparrow$ & Self-BLEU $\downarrow$ \\
\midrule
AceGPT-v2-8B-Chat        & 67.94 & 22.02 &  6.94 & 21.59 & 0.6688 & 0.7902 & 0.9409 & 0.310 \\
ALLaM-7B (base)          & 47.97 & 32.63 &  7.18 & 21.27 & 0.6796 & 0.7616 & 0.9142 & 0.017 \\
ALLaM-LoRA-No-Token      & 80.50 &  9.26 &  4.55 & 23.70 & 0.7377 & 0.9038 & 0.9881 & 0.600 \\
\textbf{ALLaM-LoRA-Token} & \textbf{84.21} & \textbf{6.21} & \textbf{4.90} & \textbf{24.80} & \textbf{0.7386} & \textbf{0.8875} & \textbf{0.9838} & \textbf{0.660} \\
Falcon-7B-Instruct       & 55.62 & 18.80 & 13.52 & 17.81 & 0.6321 & 0.7745 & 0.9073 & 0.219 \\
JAIS-13B-Chat            & 28.83 & 44.27 & 10.41 & 15.95 & 0.6581 & 0.6933 & 0.8087 & 0.350 \\
Llama-3.1-8B-Instruct    & 65.55 & 11.10 &  9.69 & 17.41 & 0.6290 & 0.6605 & 0.7957 & 0.051 \\
Qwen-2.5-7B-Instruct     & 50.12 &  7.72 & 13.52 & 19.15 & 0.6327 & 0.6606 & 0.7957 & 0.061 \\
\bottomrule
\end{tabular}%
}
\end{table*}

\subsection{Detailed Metrics}\vspace{-1ex}
Table~\ref{tab:dialect} reports full evaluation results for baselines and fine-tuned variants.  
Metrics are grouped into two categories: \textit{dialect fidelity} (Saudi\%, MSA leak, Low-conf, chrF++, BERTScore) and \textit{diversity} (distinct-2/3, Self-BLEU).  
Our LoRA models achieve the best balance between fidelity and diversity, consistently outperforming external baselines.

\subsection{Comparison with Baselines}\vspace{-1ex}
Relative to AceGPT-v2-8B-Chat, Falcon-7B-Instruct, JAIS-13B-Chat, and Llama-3.1-8B-Instruct, our fine-tuned variants yield higher Saudi correctness, improved chrF++ and BERTScore, and stronger diversity.  
The \texttt{ALLaM-LoRA-Token} variant is particularly effective, reducing MSA leakage while maintaining stable dialectal alignment.  
Both automatic and human evaluations consistently rank our models above baselines, particularly in dialect correctness.

\vspace{-4ex}

\subsection{Error Analysis}\vspace{-1ex}Residual MSA leakage appears most often in formal prompts, where training coverage is limited.  
For example:

\noindent\textbf{Prompt:} \RL{اشرح فوائد التعليم عن بُعد بالهجري} \\
\noindent\textbf{Output:} \RL{إن التعليم عن بُعد يساهم في توفير فرص التعلم للجميع...}

We also observe occasional repetition in short, high-frequency prompts:

\noindent\textbf{Prompt:} \RL{عطني مثال على مثل شعبي} \\
\noindent\textbf{Output:} \RL{المثل يقول ... المثل يقول ...} \hspace{10pt}\textit{(repetition pattern)}

Although such errors are less frequent than in external baselines, they highlight the need for more conversational and stylistically varied Saudi training data. Annotators similarly noted occasional over-formality and repetition, but these issues were significantly less frequent in the token-based model.

\vspace{-4ex}
\section{Limitations}\vspace{-2ex}Saudi-Dialect-ALLaM was trained on single-turn, synthetic, dialect-pure text and lacks explicit supervision for instruction-following or multi-turn dialogue. Consequently, the model is optimized for free-form dialectal generation rather than task-oriented dialogue or question answering, and it may drift toward formality on out-of-distribution prompts.  

Our automatic evaluation relies on the MARBERTv2 classifier’s GLF label as a proxy for Saudi usage, which can misclassify borderline cases and does not perfectly separate Hijazi from Najdi. Evaluation is also restricted to written text and short outputs; we do not assess speech, code-switching, or robustness to adversarial prompts.  

The dataset’s modest size (5.5k pairs) and limited topical coverage may introduce bias and restrict the ability to capture rarer Saudi expressions. Future work should expand the dataset with larger-scale instruction-tuning data, multi-turn conversations, broader safety screening, and a Saudi-specific dialect classifier.  

Finally, our human evaluation is limited in scope: only three annotators and 100 prompts per model were used, restricted to single-turn judgments. The use of Likert scales may also introduce subjective bias.

\section{Conclusion}
We curated a balanced synthetic corpus covering Hijazi and Najdi Arabic and used it to fine-tune ALLaM-7B-Instruct-preview with LoRA. Our experiments compared Dialect-Token and No-Token strategies, showing that explicit token conditioning significantly improves dialect correctness and reduces MSA leakage while preserving fluency and task adherence. Both automatic metrics and human evaluation by native Saudi speakers confirm the advantages of the token-based model, which consistently outperforms strong external baselines. This work highlights the value of targeted, dialect-specific resources for adapting Arabic LLMs and provides a reproducible framework for future Saudi-centric NLP research. Future extensions include scaling to multi-turn conversations, developing Saudi-specific dialect classifiers, and expanding human evaluations.

\section*{Data Availability}
The Saudi Dialect Instruction dataset used in this study is \textbf{not publicly available}. We also do \textbf{not} release any model weights or LoRA adapters. We release only training, evaluation, and inference code, configuration files (hyperparameters and random seeds), and a datasheet describing the dataset schema, topic taxonomy, cleaning rules, and aggregate statistics. These artifacts enable independent verification of our results without access to the raw training data or model weights.

\section*{Responsible Use \& Legal Considerations}
Our instruction–response pairs were synthesized via API-assisted prompting and then curated. To respect provider terms and avoid redistribution risks, we do not release the raw dataset or any fine-tuned weights/adapters derived from it. We instead release code, configs, and a datasheet (schema, cleaning rules, topic taxonomy, and aggregate statistics) so others can reproduce our pipeline on their own data. The work complies with the licenses of all referenced models and APIs.

\section*{Acknowledgments}
We thank the annotators for their contributions to the human evaluation and the Saudi NLP community for inspiring this work. This research benefited from open-source efforts, including ALLaM, MARBERT, and related Arabic NLP resources. We also acknowledge the use of AI tools for grammar checking and formatting support. Finally, we thank IEEE for providing the \texttt{IEEEtran} \LaTeX{} template.

\setcode{utf8}
\bibliographystyle{IEEEtran}
\bibliography{refs}

\begin{thebibliography}{10}
\providecommand{\url}[1]{#1}
\csname url@samestyle\endcsname
\providecommand{\newblock}{\relax}
\providecommand{\bibinfo}[2]{#2}
\providecommand{\BIBentrySTDinterwordspacing}{\spaceskip=0pt\relax}
\providecommand{\BIBentryALTinterwordstretchfactor}{4}
\providecommand{\BIBentryALTinterwordspacing}{\spaceskip=\fontdimen2\font plus
\BIBentryALTinterwordstretchfactor\fontdimen3\font minus \fontdimen4\font\relax}
\providecommand{\BIBforeignlanguage}[2]{{%
\expandafter\ifx\csname l@#1\endcsname\relax
\typeout{** WARNING: IEEEtran.bst: No hyphenation pattern has been}%
\typeout{** loaded for the language `#1'. Using the pattern for}%
\typeout{** the default language instead.}%
\else
\language=\csname l@#1\endcsname
\fi
#2}}
\providecommand{\BIBdecl}{\relax}
\BIBdecl

\bibitem{llama3modelcard}
\BIBentryALTinterwordspacing
{Meta AI}, ``Llama 3 model card,'' 2024. [Online]. Available: \url{https://github.com/meta-llama/llama3/blob/main/MODEL_CARD.md}
\BIBentrySTDinterwordspacing

\bibitem{bari2025allam}
\BIBentryALTinterwordspacing
M.~S. Bari, Y.~Alnumay, N.~A. Alzahrani, N.~M. Alotaibi, H.~A. Alyahya, S.~AlRashed, F.~A. Mirza, S.~Z. Alsubaie, H.~A. Alahmed, G.~Alabduljabbar, R.~Alkhathran, Y.~Almushayqih, R.~Alnajim, S.~Alsubaihi, M.~Al~Mansour, S.~A. Hassan, M.~Alrubaian, A.~Alammari, Z.~Alawami, A.~Al-Thubaity, A.~Abdelali, J.~Kuriakose, A.~Abujabal, N.~Al-Twairesh, A.~Alowisheq, and H.~Khan, ``{ALL}am: Large language models for arabic and english,'' in \emph{ICLR}, 2025. [Online]. Available: \url{https://openreview.net/forum?id=MscdsFVZrN}
\BIBentrySTDinterwordspacing

\bibitem{sengupta2023jais}
N.~Sengupta, S.~K. Sahu, B.~Jia, S.~Katipomu, H.~Li, F.~Koto, O.~M. Afzal, S.~Kamboj, O.~Pandit, R.~Pal, L.~Pradhan, Z.~M. Mujahid, M.~Baali, A.~F. Aji, Z.~Liu, A.~Hock, A.~Feldman, J.~Lee, A.~Jackson, P.~Nakov, T.~Baldwin, and E.~Xing, ``Jais and jais-chat: Arabic-centric foundation and instruction-tuned open generative large language models,'' 2023.

\bibitem{antoun-etal-2021-aragpt2}
\BIBentryALTinterwordspacing
W.~Antoun, F.~Baly, and H.~Hajj, ``{A}ra{GPT}2: Pre-trained transformer for {A}rabic language generation,'' in \emph{Proceedings of the Sixth Arabic Natural Language Processing Workshop}.\hskip 1em plus 0.5em minus 0.4em\relax Kyiv, Ukraine (Virtual): Association for Computational Linguistics, Apr. 2021, pp. 196--207. [Online]. Available: \url{https://www.aclweb.org/anthology/2021.wanlp-1.21}
\BIBentrySTDinterwordspacing

\bibitem{OALL-2}
A.~El~Filali, M.~ALOUI, T.~Husaain, A.~Alzubaidi, B.~E.~A. Boussaha, R.~Cojocaru, C.~Fourrier, N.~Habib, and H.~Hacid, ``Open arabic llm leaderboard 2,'' https://huggingface.co/spaces/OALL/Open-Arabic-LLM-Leaderboard, 2025.

\bibitem{abdulmageed2021_marbert}
\BIBentryALTinterwordspacing
M.~Abdul-Mageed, A.~Elmadany, and E.~M.~B. Nagoudi, ``Arbert \& marbert: Deep bidirectional transformers for arabic,'' in \emph{ACL}, 2021. [Online]. Available: \url{https://aclanthology.org/2021.acl-long.551/}
\BIBentrySTDinterwordspacing

\bibitem{ibrahimamin_marbertv2_arabic_written_dialect_classifier}
\BIBentryALTinterwordspacing
I.~Amin, ``Marbertv2 arabic written dialect classifier,'' 2025. [Online]. Available: \url{https://huggingface.co/IbrahimAmin/marbertv2-arabic-written-dialect-classifier}
\BIBentrySTDinterwordspacing

\bibitem{falcon40b}
\BIBentryALTinterwordspacing
E.~Almazrouei, H.~Alobeidli, A.~Alshamsi, A.~Cappelli, R.~Cojocaru, M.~Debbah, E.~Goffinet, D.~Heslow, J.~Launay, Q.~Malartic, B.~Noune, B.~Pannier, and G.~Penedo, ``{Falcon-40B}: An open large language model with state-of-the-art performance,'' 2023. [Online]. Available: \url{https://huggingface.co/tiiuae/falcon-7b-instruct}
\BIBentrySTDinterwordspacing

\bibitem{falcon-arabic}
\BIBentryALTinterwordspacing
{Falcon-LLM Team}, ``Falcon-arabic: A breakthrough in arabic language models,'' May 2025. [Online]. Available: \url{https://falcon-lm.github.io/blog/falcon-arabic}
\BIBentrySTDinterwordspacing

\bibitem{qwen2.5}
\BIBentryALTinterwordspacing
{Qwen Team}, ``Qwen2.5: A party of foundation models,'' September 2024. [Online]. Available: \url{https://qwenlm.github.io/blog/qwen2.5/}
\BIBentrySTDinterwordspacing

\bibitem{liang2024alignment}
\BIBentryALTinterwordspacing
J.~Liang, Z.~Cai, J.~Zhu, H.~Huang, K.~Zong, B.~An, M.~Alharthi, J.~He, L.~Zhang, H.~Li, B.~Wang, and J.~Xu, ``Alignment at pre-training! towards native alignment for arabic llms,'' in \emph{NeurIPS}, 2024. [Online]. Available: \url{https://openreview.net/forum?id=woRFmNJiLp}
\BIBentrySTDinterwordspacing

\bibitem{humaine_huwaimain2025}
\BIBentryALTinterwordspacing
{Asharq Al-Awsat}, ``Humane launches allam, the first arabic artificial intelligence model from saudi arabia, august 2025,'' 2025. [Online]. Available: \url{https://aawsat.com/%D8%A7%D9%84%D8%A7%D9%82%D8%AA%D8%B5%D8%A7%D8%AF/5174385-%D9%87%D9%8A%D9%88%D9%85%D8%A7%D9%8A%D9%86-%D8%A5%D8%B7%D9%84%D8%A7%D9%82-%D8%B9%D9%84%D9%91%D8%A7%D9%85-%D8%A3%D9%88%D9%84-%D9%86%D9%85%D9%88%D8%B0%D8%AC-%D8%B0%D9%83%D8%A7%D8%A1-%D8%A7%D8%B5%D8%B7%D9%86%D8%A7%D8%B9%D9%8A-%D8%B9%D8%B1%D8%A8%D9%8A-%D9%85%D9%86-%D8%A7%D9%84%D8%B3%D8%B9%D9%88%D8%AF%D9%8A%D8%A9-%D8%A3%D9%88%D8%A7%D8%AE%D8%B1-%D8%A3%D8%BA%D8%B3%D8%B7%D8%B3}
\BIBentrySTDinterwordspacing

\bibitem{qarah2024saudibert}
F.~Qarah, ``Saudibert: A large language model pretrained on saudi dialect corpora,'' \emph{arXiv preprint arXiv:2405.06239}, 2024.

\bibitem{abdul-mageed-etal-2021-arbert}
\BIBentryALTinterwordspacing
M.~Abdul-Mageed, A.~Elmadany, and E.~M.~B. Nagoudi, ``Arbert \& marbert: Deep bidirectional transformers for arabic,'' in \emph{Proceedings of the 59th Annual Meeting of the Association for Computational Linguistics and the 11th International Joint Conference on Natural Language Processing (Volume 1: Long Papers)}.\hskip 1em plus 0.5em minus 0.4em\relax misc: Association for Computational Linguistics, Aug. 2021, pp. 7088--7105. [Online]. Available: \url{https://aclanthology.org/2021.acl-long.551}
\BIBentrySTDinterwordspacing

\bibitem{gemmar_arxiv_2024}
\BIBentryALTinterwordspacing
H.~Chouikhi, M.~Aloui, C.~Ben~Hammou, G.~Chaabane, H.~Kchaou, and C.~Dhaouadi, ``Gemmar: Enhancing llms through arabic instruction-tuning,'' \emph{arXiv preprint arXiv:2407.02147}, 2024. [Online]. Available: \url{https://arxiv.org/abs/2407.02147}
\BIBentrySTDinterwordspacing

\bibitem{chouikhi2024}
------, ``Llamar \& gemmar: Enhancing llms through arabic instruction-tuning,'' 2024.

\bibitem{aburayyash2025saudial}
\BIBentryALTinterwordspacing
H.~Abu-Rayyash and N.~Alanazi, ``Saudial: The saudi arabic dialects game localization dataset,'' 2025. [Online]. Available: \url{https://data.mendeley.com/datasets/mzdwkb2t6d/2}
\BIBentrySTDinterwordspacing

\bibitem{alahmari-2025-sadslyc}
\BIBentryALTinterwordspacing
S.~S. Alahmari, ``{SADSL}y{C}: A corpus for saudi {A}rabian multi-dialect identification through song lyrics,'' in \emph{Proceedings of the 4th Workshop on Arabic Corpus Linguistics (WACL-4)}, S.~Ezzini, H.~Alami, I.~Berrada, A.~Benlahbib, A.~El~Mahdaouy, S.~Lamsiyah, H.~Derrouz, A.~Haddad, M.~Jarrar, M.~El-Haj, R.~Mitkov, and P.~Rayson, Eds.\hskip 1em plus 0.5em minus 0.4em\relax Abu Dhabi, UAE: Association for Computational Linguistics, Jan. 2025, pp. 38--43. [Online]. Available: \url{https://aclanthology.org/2025.wacl-1.4/}
\BIBentrySTDinterwordspacing

\bibitem{sada2022}
\BIBentryALTinterwordspacing
{National Center for Artificial Intelligence (SDAIA)} and {Saudi Broadcasting Authority}, ``Sada: Saudi audio dataset for arabic,'' 2022, 667 hours of Saudi Arabic audio with transcripts across 57 TV programs. [Online]. Available: \url{https://www.kaggle.com/datasets/sdaiancai/sada2022}
\BIBentrySTDinterwordspacing

\bibitem{vaswani2017attention}
\BIBentryALTinterwordspacing
A.~Vaswani, N.~Shazeer, N.~Parmar, J.~Uszkoreit, L.~Jones, A.~N. Gomez, {\L}.~Kaiser, and I.~Polosukhin, ``Attention is all you need,'' in \emph{Advances in Neural Information Processing Systems (NeurIPS)}, vol.~30, 2017, pp. 5998--6008. [Online]. Available: \url{https://arxiv.org/abs/1706.03762}
\BIBentrySTDinterwordspacing

\bibitem{hu2021lora}
\BIBentryALTinterwordspacing
E.~J. Hu, Y.~Shen, P.~Wallis, Z.~Allen-Zhu, Y.~Li, L.~Wang, and W.~Wang, ``Lora: Low-rank adaptation of large language models,'' \emph{arXiv preprint arXiv:2106.09685}, 2021. [Online]. Available: \url{https://doi.org/10.48550/arXiv.2106.09685}
\BIBentrySTDinterwordspacing

\bibitem{holtzman2020curious}
\BIBentryALTinterwordspacing
A.~Holtzman, J.~Buys, L.~Du, M.~Forbes, and Y.~Choi, ``The curious case of neural text degeneration,'' in \emph{International Conference on Learning Representations (ICLR)}, 2020. [Online]. Available: \url{https://arxiv.org/abs/1904.09751}
\BIBentrySTDinterwordspacing

\bibitem{popovic2017chrfpp}
\BIBentryALTinterwordspacing
M.~Popovi{\'c}, ``chrf++: words helping character n-grams,'' in \emph{Proceedings of the Second Conference on Machine Translation (WMT)}.\hskip 1em plus 0.5em minus 0.4em\relax Copenhagen, Denmark: Association for Computational Linguistics, Sep. 2017, pp. 612--618. [Online]. Available: \url{https://aclanthology.org/W17-4770/}
\BIBentrySTDinterwordspacing

\bibitem{zhang2020bertscore}
\BIBentryALTinterwordspacing
T.~Zhang, V.~Kishore, F.~Wu, K.~Q. Weinberger, and Y.~Artzi, ``Bertscore: Evaluating text generation with bert,'' in \emph{International Conference on Learning Representations (ICLR)}, 2020. [Online]. Available: \url{https://arxiv.org/abs/1904.09675}
\BIBentrySTDinterwordspacing

\bibitem{li2016diversity}
\BIBentryALTinterwordspacing
J.~Li, M.~Galley, C.~Brockett, J.~Gao, and B.~Dolan, ``A diversity-promoting objective function for neural conversation models,'' in \emph{Proceedings of the 2016 Conference of the North American Chapter of the Association for Computational Linguistics: Human Language Technologies (NAACL-HLT)}.\hskip 1em plus 0.5em minus 0.4em\relax San Diego, California: Association for Computational Linguistics, 2016, pp. 110--119. [Online]. Available: \url{https://aclanthology.org/N16-1014/}
\BIBentrySTDinterwordspacing

\bibitem{zhu2018texygen}
\BIBentryALTinterwordspacing
Y.~Zhu, S.~Lu, L.~Zheng, J.~Guo, W.~Zhang, J.~Wang, and Y.~Yu, ``Texygen: A benchmarking platform for text generation models,'' \emph{arXiv preprint arXiv:1802.01886}, 2018. [Online]. Available: \url{https://arxiv.org/abs/1802.01886}
\BIBentrySTDinterwordspacing

\end{thebibliography}

\end{document}